\documentclass[twocolumn]{article}

\usepackage{arxiv}
\usepackage{amsmath}
\usepackage[utf8]{inputenc} 
\usepackage[T1]{fontenc}    
\usepackage{hyperref}       
\usepackage{url}            
\usepackage{booktabs}       
\usepackage{amsfonts}       
\usepackage{nicefrac}       
\usepackage{microtype}      
\usepackage{cleveref}       
\usepackage{lipsum}         
\usepackage{graphicx}
\usepackage[numbers]{natbib}
\usepackage{doi}

\title{Forecasting Ferry Passenger Flow using Long-Short Term Memory Neural Networks}

\date{}

\newif\ifuniqueAffiliation
\uniqueAffiliationtrue

\ifuniqueAffiliation 
\author{ \href{https://orcid.org/0009-0008-4770-3623}{\includegraphics[scale=0.06]{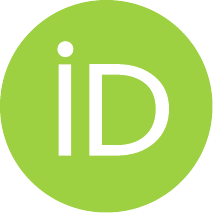}\hspace{1mm}Daniel F.~Fesalbon}\thanks{Use footnote for providing further
		information about author (webpage, alternative
		address)---\emph{not} for acknowledging funding agencies.} \\
}
\else
\usepackage{authblk}

\newbox{\orcid}\sbox{\orcid}{\includegraphics[scale=0.06]{orcid.pdf}} 
\author[1]{%
	\href{https://orcid.org/0009-0008-4770-3623}{\usebox{\orcid}\hspace{1mm}Daniel F.~Fesalbon\thanks{\texttt{danielffesalbon@iskolarngbayan.pup.edu.ph}}}%
}
\affil[1]{Polytechnic University of the Philippines}
\fi

\renewcommand{\headeright}{IJSRED – Volume 7 Issue 3, May-June 2024}
\renewcommand{\undertitle}{A Research Article}
\renewcommand{\shorttitle}{}
\hypersetup{
colorlinks=false,
pdfborder={0 0 0},
pdfborderstyle={/S/U/W 0},
pdftitle={Forecasting Ferry Passenger Flow using Long-Short Term Memory Neural Networks},
pdfsubject={},
pdfauthor={Daniel F.~Fesalbon},
pdfkeywords={Time Series Analysis, Ferry Passenger, Passenger Traffic Forecasting, Neural Networks},
}

\begin{document}
\twocolumn[ 
  \begin{@twocolumnfalse} 
  
\maketitle

\begin{abstract}
With recent studies related to Neural Networks being used on different forecasting and time series investigations, this study aims to expand these contexts to ferry passenger traffic. The primary objective of the study is to investigate and evaluate an LSTM-based Neural Networks’ capability to forecast ferry passengers of two ports in the Philippines. The proposed model’s fitting and evaluation of the passenger flow forecasting of the two ports is based on monthly passenger traffic from 2016 to 2022 data that was acquired from the Philippine Ports Authority (PPA). This work uses Mean Absolute Percentage Error (MAPE) as its primary metric to evaluate the model’s forecasting capability. The proposed LSTM-based Neural Networks model achieved 72\% forecasting accuracy to the Batangas port ferry passenger data and 74\% forecasting accuracy to the Mindoro port ferry passenger data. Using Keras and Scikit-learn Python libraries, this work concludes a reasonable forecasting performance of the presented LSTM model. Aside from these notable findings, this study also recommends further investigation and studies on employing other statistical, machine learning, and deep learning methods on forecasting ferry passenger flows.
\end{abstract}

\keywords{Time Series Analysis \and Ferry Passenger\and Passenger Traffic Forecasting \and Neural Networks}
\vspace{0.35cm}
  \end{@twocolumnfalse} 
] 

\section{Introduction}
There are 7,107 islands that compose the Philippines. Its geographical settings depend mostly on marine resources from its coastline and are strongly tied to maritime territories in terms of its economic and social activities \citep{angelesfa2015}. In the year 2022, the Philippine Ports Authority (PPA) reported an annual total of around 59 million passenger traffic. The majority or 45.7\% of the traffic from the 2022 report was from the Visayas region. Being an island country, ferry transportation and services in the Philippines were a major means of passage. During peak seasons it is usual for ports and different ferry terminals in the Philippines to experience heavy passenger traffic.

Time-series is a sequential type of data that is composed of observation points or data points that are ordered chronologically. Promising methods or techniques for time-series forecasting can be divided into two approaches such as statistical and machine learning (ML) methods. Methods that are considered to be statistical-based were Autoregressive (AR) models and Moving Average (MA) analysis. Artificial Neural Networks (ANN) are considered in deep learning (DL) which falls in the machine learning category. It is an ML model or program that mimics the function of biological neurons. ANN is composed of connected units or nodes called artificial neurons. Since its discovery, it has been widely used and investigated in different ML applications such as classification, forecasting, regression, anomaly detection, etc.

In handling passenger traffic, strategic and tactical decisions of an organization or management depend on an accurate forecast of passenger traffic flows and an accurate estimation can optimize an organization’s future financial planning \cite{ahmadta2021}. This study focuses on investigating and using a specific type of ANN called Long-Short Term Memory (LSTM) Neural Networks and evaluating its capacity for forecasting ferry passenger traffic at Calapan Port and Batangas Port in the Philippines.

\section{Related Studies}

Time-series data forecasting using an LSTM-based Neural Networks approach has been applied to different cases aside from ferry passenger traffic. There are numerous studies that employed LSTM Neural Networks in forecasting sequential data such as stock price prediction, air passenger traffic, temperature, wind speed, etc. One example of a study was from Feng et al. \citep{fengf2023}. Their study is about forecasting multi-level rail transit hub passenger flow. They investigate a multi-task learning model to forecast trunk railway, intercity railway, and subway passenger flow and compare it to other forecasting methods/models which includes a Fully Connected LSTM (FC-LSTM) Neural Networks model. Other methods and/or models along with the FC-LSTM and the multi-task learning model where Historical Average (HA), Spatiotemporal Graph Convolutional Network (SGCN), Gated Recurrent Unit (GRU), Support Vector Regression (SVR), Diffusion Convolutional Recurrent Neural Network (DCRNN). Among the highlight models, the multi-task learning model, HA, FC-LSTM, and SGCN, the multi-task learning model achieves the best performance. The FC-LSTM follows right after the multi-task learning model with 82.23 \% accuracy.

Another study that uses LSTM-based Neural Networks to forecast sequential data was from Riyadi et al \citep{riyadiw2023}. They study whether an LSTM and a Combined Convolutional Neural Network (CNN) and LSTM (CNN-LSTM) offer more accurate predictions for airport traffic during the COVID-19 pandemic from March to December 2020. The data that they used were from countries USA, Canada, Chile, and Australia where they created an average airport baseline on a daily basis given the imbalanced nature of the airport dataset across said countries. They split the data with 80\% to training and 20\% to testing. They investigate the 2 models on different optimization parameters such as RMSProp, Stochastic Gradient Descent (SGD), Adam, Nadam, and Adamax. They implemented a 2-layer LSTM on the base LSTM. The ones that achieved the highest accuracy from the two models were the ones with used the Adamax optimizer. The Adamax-optimized LSTM (base) achieves 90.68\% accuracy while the Adamax-optimized CNN-LSTM achieves 90.4\% accuracy. Their study also recommends to conduct further research to address the data imbalance issue.

A similar study that proposes 2-layer LSTM Recurrent Neural Networks (LSTM-RNN) for forecasting airline passenger flow was from Gupta et al \citep{guptavs2019}. They compare the proposed 2-layer LSTM-RNN to different existing studies that use Single Layer Multiple Inputs LSTM, a study with Holt-Winters, and a study with a method that combines Back-propagation Neural Networks and Genetic Algorithm. Their proposed 2-layer LSTM-RNN achieves 89.2\% accuracy on the training data and 91.21\% on the test data. They also recommend incorporating convolution layers in the architecture and trying to make the model even deeper using multiple LSTM layers.

One study that combines an LSTM with another technique was from Hou et al \citep{houz2022}. They combine LSTM and TCN (Time Convolution Network) to forecast passenger flow in three urban rail stations. They couple external factors such as date attributes, weather conditions, and air quality, to improve the overall prediction and compare a single base LSTM to the TCN-LSTM combination. The TCN-LSTM model achieves better prediction achieving 87.84\% accuracy compared to the base LSTM which achieves 79.02\% accuracy. Their findings concluded that the TCN-LSTM has better prediction accuracy and data generalization ability.

Another study incorporating other techniques to an LSTM model in a forecasting investigation was from Farahani et al \citep{farahanim2020}. They present a novel hybrid deep learning model based on a Variational Long Short-Term Memory Encoder (VLSTM-E) to forecast the short-term traffic flow of two points between two stations. They compare the VLSTM-E to other models/methods such as Multiple Convolutional Neural Network for Multivariate (MCNNM), Stacked Autoencoders (SAEs), and a base LSTM. The VLSTM-E achieves 90.771\% accuracy. From these existing studies, it can be seen that an LSTM has a remarkable performance in forecasting time-series or sequential data especially on passenger traffic. Aside from that, its capability to be incorporated or combined with other techniques and methods is promising. For this study, the proposed LSTM-RNN will be a 2-layer LSTM with an Adam optimizer.

\section{Methods}

\subsection{Long-Short Term Memory Recurrent Neural Networks}

LSTM is based on another architecture of NN called Recurrent Neural Networks (RNN). RNNs are a superset of feed-forward neural networks, augmented with the ability to pass information across time steps \citep{liptonzc2015}, and can process sequential or time-series data. RNNs are known to have gradient descent issues \citep{bengioy1994} and LSTM is a type of RNN that was developed to address the vanishing gradients problem \citep{sherstinskya2020}.

\begin{figure}[h]
	\caption{Illustrates the structure of the LSTM memory cell \citep{andréassond2020} consisting of four elements such as forget gate (\textit{f\textsubscript{t}}), input gate (\textit{i\textsubscript{t}}), reconnection neuron ($\tilde{C\textsubscript{t}}$), and the output gate (\textit{o\textsubscript{t}}).}
	\centering
	\includegraphics[width=7 cm]{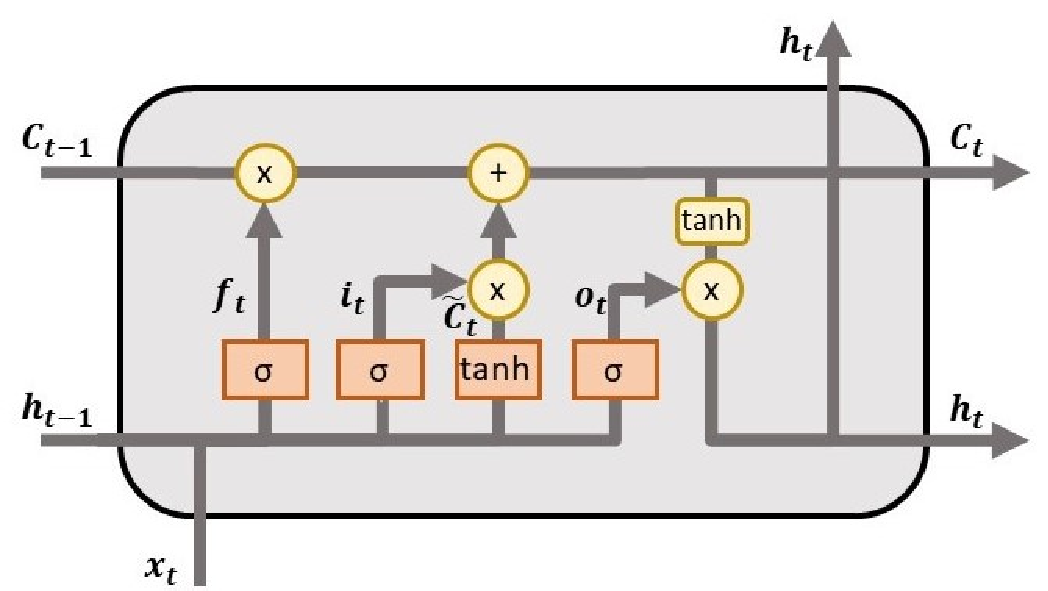}
	\label{fig5:fig5}
\end{figure}

Derivation of components showed in the LSTM Memory cell are the following:
\begin{align}
&i_t=\sigma (W_i*[h_{t-1},x_t ]+b_i) \\
&f_t=\sigma(W_f*[h_{t-1}x_t ]+b_f) \\
&\tilde{C} _t=tanh(W_C*[h_{t-1},x_t ]+b_C) \\
&C_t=f_t*C_{t-1}+i_t*\tilde{C}_t \\
&o_t=\sigma (W_o*[h_{t-1},x_t ]+b_o)\\
&h_{t}=o_{t}\ast tanh(C_{t })\\
&\sigma_x=\frac{1}{1+ e^x}\\
&tanh_x= \frac{ e^{x}-e^{-x}}{ e^{x}+e^{-x}} 
\end{align}

(\textit{W}) denotes the weight matrix, (\textit{b})  represents the bias, (\textit{X\textsubscript{t}}) represents the input at the current time, and (\textit{h\textsubscript{t-1}}) as the input from the previous timestep. Equation 1 described the input gate calculation that determines the values to be updated. Followed by Equation 3 which is the updated gate creating a vector of new potential memory. Equation 2 shows the forget gate operation. In Equation 4, the two previous equations are combined and merged with the previous memory in order to get the new memory. Lastly, Equation 5 decides the output, which is multiplied with the current memory in Equation 6. Equation 7 shows the sigmoid ($\sigma$) activation function calculation and Equation 8 for the (\textit{tanh})  activation function \citep{andréassond2020}.

\subsection{Data Acquisition and Processing}

The study uses data that is obtained from the Philippine Ports Authority (PPA). The data consists of quarterly and monthly ferry passengers of different ferry terminals in the Philippines from 2015 up to 2022. The data is a chronologically ordered one-dimensional sequence which will be formatted into an \textit{X} and \textit{Y} row or tabular format. In an iterative and incremental method, row data will be consisting of four data points from the sequence. \textit{X} will be having first three data points with \textit{Y} always taking the fourth data point being the target value or the one to be forecasted. See the following example:

Original dataset structure/format:

\[ [t\textsubscript{1}, t\textsubscript{2}, t\textsubscript{3}, t\textsubscript{4}, t\textsubscript{5}, t\textsubscript{6}, t\textsubscript{7}, \ldots t\textsubscript{n}]\]
\[\downarrow\]
\indent Converted tabular dataset structure/format: 

\[X\textsubscript{1} = [t\textsubscript{1}, t\textsubscript{2}, t\textsubscript{3}],  Y\textsubscript{1} = t\textsubscript{4} \]
\[X\textsubscript{2} = [t\textsubscript{2}, t\textsubscript{3}, t\textsubscript{4}],  Y\textsubscript{2} = t\textsubscript{5} \]
\[X\textsubscript{3} = [t\textsubscript{3}, t\textsubscript{4}, t\textsubscript{5}],  Y\textsubscript{3} = t\textsubscript{6} \]
\[X\textsubscript{4} = [t\textsubscript{4}, t\textsubscript{5}, t\textsubscript{6}],  Y\textsubscript{4} = t\textsubscript{7}\]

\subsection{Evaluation Metrics}

To evaluate the LSTM-RNN model, the study uses Root Mean Squared Error (RMSE), Mean Absolute Error (MAE), Mean Absolute Percentage Error (MAPE), and Mean Squared Error (MSE) as criteria with the MAPE as the primary metric.

\begin{figure}[h]
	\caption{LSTM Model's Forecast/Prediction on Port Mindoro Monthly Passenger Test Sets}
	\centering
	\includegraphics[width=7 cm]{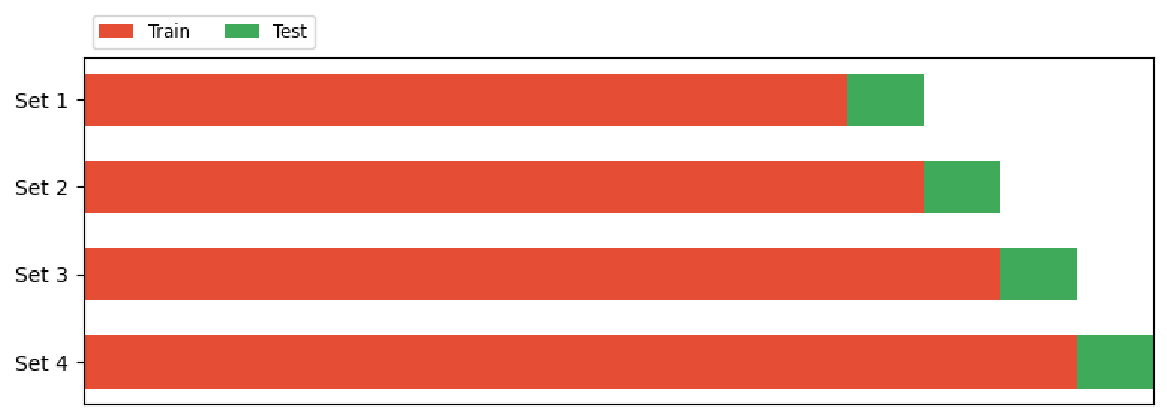}
	\label{fig4:fig4}
\end{figure}

Similar to Al-Sultan et al. \citep{ahmadta2021}, the study splits the formatted dataset into multiple train-test splits to obtain precise forecasting accuracy. The model is trained or fitted into a training set and will forecast newly introduced data which is the test set. For this study, four train-test splits were used as shown in Figure~\ref{fig4:fig4}. Test data consists of six samples and the initial train data consists of 60 data points after every training-testing iteration, the test data of the completed iteration is added to the train data on the next iteration.

\section{Results}

\subsection{LSTM Model Specification}

The study proposed a two-layer LSTM Neural Network model consisting of the input layer and one hidden layer. Specifications of the proposed LSTM model are displayed in Table~\ref{tab:table1}. The proposed model also includes an Adam optimization. The work conducted was based from Python programming machine learning libraries such as Keras \citep{cholletf2015} and Scikit-learn \citep{pedregosaf2011}.

\begin{table}[h]
	\caption{LSTM Model Specifications}
	\centering
	\begin{tabular}{llll}
		\toprule
\textbf{Layer}					& \textbf{Type}	& \textbf{Output Shape}	& \textbf{Param \#}\\
\midrule
1\textsuperscript{st} Layer (input)		& LSTM		& (None, 1, 12)			& 768\\
1\textsuperscript{st} Layer (input)		& Dropout		& (None, 1, 12)			& 0\\
2\textsuperscript{nd} Layer (hidden)		& LSTM\_1		& (None, 12)			& 1200\\
2\textsuperscript{nd} Layer (hidden)		& Dropout\_1	& (None, 12)			& 0\\
Output Layer					& Dense		& (None, 1)				& 13\\
\bottomrule
	\end{tabular}
	\label{tab:table1}
\end{table}

\subsection{Ferry Passenger Data}

\begin{figure}[h]
	\caption{Batangas Port Monthly Passenger Traffic from 2016 up to 2022}
	\centering
	\includegraphics[width=7 cm]{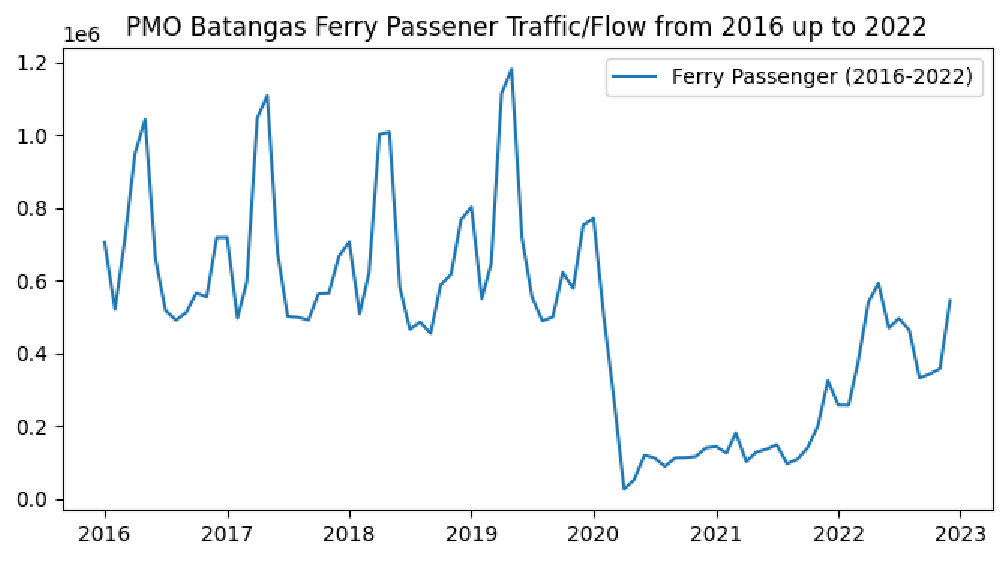}	
	\label{fig0:fig0}

	\caption{Mindoro Port Monthly Passenger Traffic from 2016 up to 2022}
	\centering
	\includegraphics[width=7 cm]{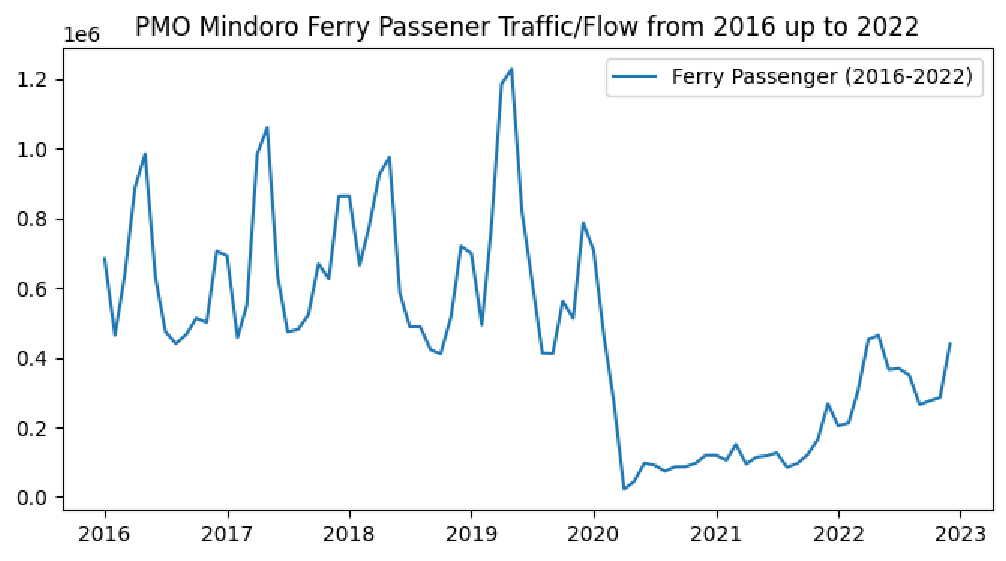}
	\label{fig1:fig1}
\end{figure}

Figure~\ref{fig0:fig0} and~\ref{fig1:fig1} display the monthly passenger traffic of the two ports investigated in this study. Passenger traffic data per port consists of 84 observation points from 2016 and 2022.

\subsection{Model Training and Forecasting Performance}

The model was trained with 1200 epochs and 32 batch sizes. The model achieved both reasonable forecasting performance on Batangas Port and Mindoro Port. Attaining 0.28 Average MAPE on the Batangas Ferry Passenger Forecast and 0.26 Average MAPE on the Mindoro Ferry Passenger Forecast. Aside from MAPE, Table~\ref{tab:table2} and~\ref{tab:table3} also display MAE, MSE, and RMSE.

\begin{table}[!h]
	\caption{Port Batangas Forecasting Evaluation}
	\centering
	\resizebox{8cm}{!}{%
	\begin{tabular}{lllll}
\toprule
\textbf{Set/Split}	& \textbf{MAE}	& \textbf{MAPE}	& \textbf{MSE}	& \textbf{RMSE}\\
\midrule
1		& 30844.05			& 0.22		& 1371556170.66			& 37034.53\\
2		& 60979.45			& 0.30		& 7076462372.00			& 84121.71\\
3		& 138742.10		& 0.37		& 25484350455.31			& 159638.19\\
4		& 97779.64			& 0.24		& 13566002842.65			& 116473.19\\
\bottomrule
Average	& 82086.31			& \textbf{0.28}		& 11874592960.15			& 99316.90\\

	\end{tabular}
	}
	\label{tab:table2}
\end{table}
\begin{table}[!h]
	\caption{Port Mindoro Forecasting Evaluation}
	\centering
	\resizebox{8cm}{!}{%
	\begin{tabular}{lllll}
\toprule
\textbf{Set/Split}	& \textbf{MAE}	& \textbf{MAPE}	& \textbf{MSE}	& \textbf{RMSE}\\
\midrule
1		& 18590.93			& 0.15		& 561271231.67			& 23691.16\\
2		& 49499.48			& 0.27		& 5546986605.38			& 74478.09\\
3		& 141880.35		& 0.41		& 25795098231.98			& 160608.52\\
4		& 70390.26			& 0.21		& 6962392677.66			& 83440.95\\
\bottomrule
Average	& 70090.25			& \textbf{0.26}		& 9716437186.67			& 85554.68\\

	\end{tabular}
	}
	\label{tab:table3}
\end{table}

\begin{figure}[h]
	\caption{LSTM Model's Forecast/Prediction on Port Batangas Monthly Passenger Test Sets}
	\centering
	\includegraphics[width=7 cm]{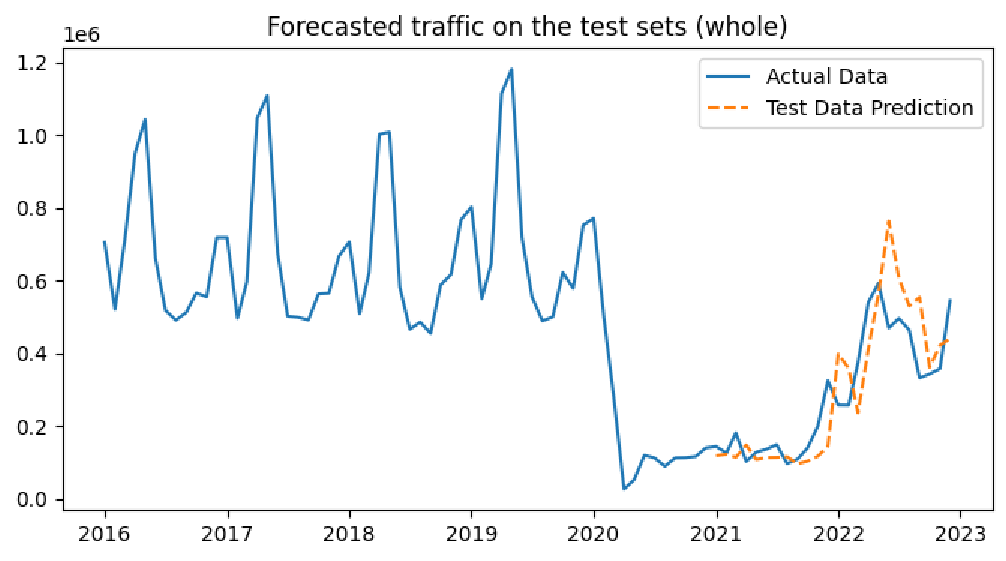}	
	\label{fig2:fig2}

	\caption{LSTM Model's Forecast/Prediction on Port Mindoro Monthly Passenger Test Sets}
	\centering
	\includegraphics[width=7 cm]{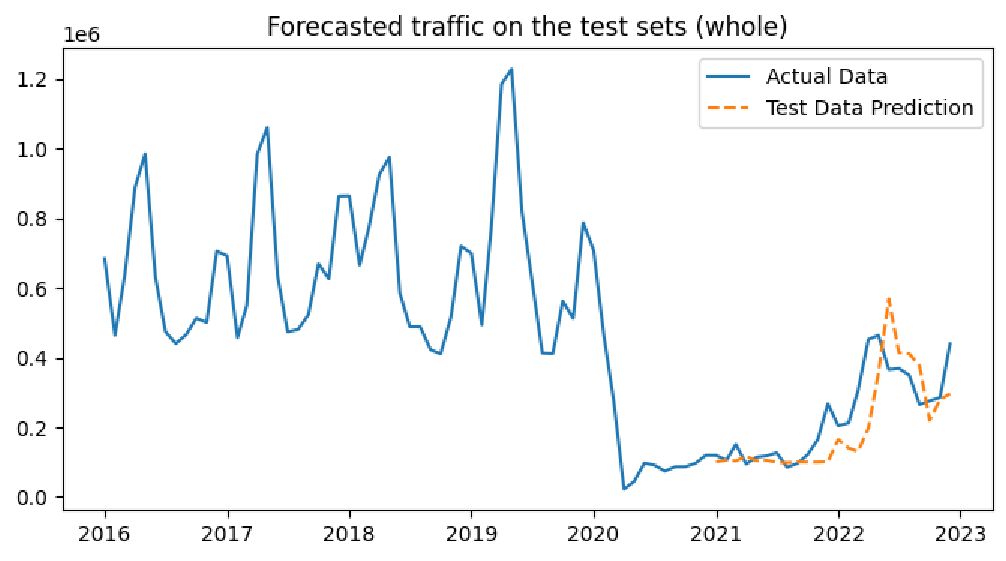}
	\label{fig3:fig3}
\end{figure}

Figure~\ref{fig2:fig2} and~\ref{fig3:fig3} illustrates the model’s forecasted data comparison to the actual observed data.

\section{Conclusions}

Ferry Passenger Traffic from two Ports in the Philippines was collected and processed for time-series forecasting in this work. Time-series forecasting can provide an edge in terms of decision-making in handling and managing transportation and passenger traffic. Analysis of these sequential data can also offer the capability to understand patterns based on past observations and forecast possible future trends and behavior. Using an LSTM-based Neural Network, this study investigates the ability of a deep learning model to forecast ferry passenger traffic. Batangas Port and Mindoro Port ferry passenger traffic data was acquired and employed in a time-series forecasting study using the presented LSTM-RNN model. This work follows the MAPE values interpretation based on Asrah et al. \citep{asrahnm2018}'s work.

\begin{table}[!h]
	\caption{Interpretation of Typical MAPE Values\citep{asrahnm2018}}
	\centering
	\resizebox{6cm}{!}{%
	\begin{tabular}{ll}
\toprule
\textbf{MAPE Values}	& \textbf{Interpretation}	\\
\midrule
< 0.10			& Highly accurate forecasting\\
0.10 - 0.20	& Good forecasting\\
0.20 - 0.50	& Reasonable forecasting\\
> 0.50			& Inaccurate forecasting\\
\bottomrule
	\end{tabular}
	\label{tab:table4}
	}
\end{table}

Displaying a 72\% average accuracy (0.28 average MAPE) on forecasting Batangas Port Passenger ferry passenger traffic and 74\% average accuracy (0.26 average MAPE) on forecasting Mindoro Port ferry passenger traffic, the presented LSTM-RNN model achieves a reasonable forecasting performance. Although the study concludes with a good and reasonable finding, this work also recommends further studies by investigating other ferry terminal traffic in the Philippines. Another possible work advancement from this study is to compare the LSTM-RNN model to other forecasting techniques and methods. The proponent believes further optimization of the LSTM-based Neural Network model such as testing other optimization techniques and a combination of other methods can achieve better and more precise accuracy for forecasting ferry passenger traffic.

\bibliographystyle{unsrtnat}


\end{document}